\documentclass{article}
\usepackage{ijcai25}

\usepackage{times}
\usepackage{soul}
\usepackage{url}
\usepackage[hidelinks]{hyperref}
\usepackage[utf8]{inputenc}
\usepackage[small]{caption}
\usepackage{graphicx}
\usepackage{amsmath}
\usepackage{amsthm}
\usepackage{booktabs}
\usepackage{xcolor}
\usepackage{algpseudocode}
\usepackage{algorithm}
\usepackage{algorithmicx}
\usepackage[switch]{lineno}
\usepackage{tikz}
\usepackage{comment}
\usepackage{tabularx}
\usepackage[inline]{enumitem}
\usetikzlibrary{fit, calc}
\newcolumntype{Z}{>{\raggedright\let\newline\\\arraybackslash\hspace{0pt}}X}
\newcommand{\tikzmk}[1]{\tikz[overlay,remember picture,baseline] \node [anchor=base] (#1) {};}%
\newcommand{\boxit}[1]{\tikz[remember picture,overlay]{\node[yshift=-5pt,fill=#1,opacity=.35,fit={($(left|-A)+(2pt,-2pt)$)($(left|-B)+(1.05\linewidth,.5\baselineskip)$)}] {
};}\ignorespaces}
\colorlet{mypink}{red!40}
\colorlet{myblue}{cyan!60}
\definecolor{boxblue}{HTML}{88ccee}
\definecolor{boxteal}{HTML}{79bdb2}
\definecolor{boxgreen}{HTML}{609271}

\definecolor{newboxyellow}{HTML}{edf8b1}
\definecolor{newboxgreen}{HTML}{7fcdbb}
\definecolor{newboxblue}{HTML}{2c7fb8}
\newcommand{\colorline}[1]{
\setlength{\fboxsep}{1pt}\hspace{-20pt}\colorbox{newboxblue!35}{\makebox[1.053\linewidth][l]{\hspace{5pt}1: \hspace{2pt}#1}}
 }

\usepackage{mathtools}
\newcommand{\defeq}{\vcentcolon=}

\title{Multi-objective methods in Federated Learning: A survey and taxonomy}

\author{
Maria Hartmann$^{1,*}$
\and
Gr\'{e}goire Danoy$^{1, 2}$\And
Pascal Bouvry$^{2}$
\affiliations
$^1$SnT, University of Luxembourg\\
$^2$FSTM/DCS, University of Luxembourg\\
$^*$Corresponding author
\emails
\{maria.hartmann, gregoire.danoy, pascal.bouvry\}@uni.lu,
}

\begin{document}

\maketitle

\begin{abstract}
The Federated Learning paradigm facilitates effective distributed machine learning in settings where training data is decentralized across multiple clients. As the popularity of the strategy grows, increasingly complex real-world problems emerge, many of which require balancing conflicting demands such as fairness, utility, and resource consumption.
Recent works have begun to recognise the use of a multi-objective perspective in answer to this challenge.
However, this novel approach of combining federated methods with multi-objective optimisation has never been discussed in the broader context of both fields. In this work, we offer a first clear and systematic overview of the different ways the two fields can be integrated.
We propose a first taxonomy on the use of multi-objective methods in connection with Federated Learning, providing a targeted survey of the state-of-the-art and proposing unambiguous labels to categorise contributions. Given the developing nature of this field, our taxonomy is designed to provide a solid basis for further research, capturing existing works while anticipating future additions. Finally, we outline open challenges and possible directions for further research. 
\end{abstract}
\section{Introduction}
\label{sec:introduction}
The Federated Learning (FL) paradigm allows the training of machine learning models in the difficult setting where training data is distributed and  compartmentalised. Instead of centralising available data, FL performs local training in distribution, with the resulting local models aggregated periodically across participants. Though originally designed to mitigate privacy concerns, the method has also shown great success in other use cases, including communication-restricted settings such as drone networks~\cite{brik_federated_2020} or computationally costly settings such as the tuning of large language models~\cite{che_federated_2023}. 

However, as Federated Learning is being adopted for increasingly diverse applications and real-world use cases, new challenges are emerging, many linked to the need to balance different conflicting requirements: 
\begin{enumerate*}[label=(\roman*)]\item Heterogeneity between participants caused by data imbalances or differing hardware capabilities can lead to divergent local models that cannot easily be aggregated without loss of model utility~\cite{karimireddy_scaffold_2019}. Designing mitigation strategies for this raises the problem of fairness -- the choice between sacrificing the performance of some individual clients or that of the global model. \item The cost of FL in terms of communication and computation resources scales with the size of the model and the number of update messages; yet reducing either may come at the cost of decreasing model utility~\cite{zhu_federated_2021}. 
\item Strategies for mitigating privacy leakage, the problem of exposing confidential information to potential attackers through client updates, may degrade other aspects of the federated system in turn. %
For example, adding noise to client updates may obscure sensitive information effectively, but reduce model performance as well~\cite{geng_multiobjective_2024}.
\end{enumerate*}

\par All these scenarios can be modelled as multi-objective problems, with each problem-specific performance metric represented as a separate objective. Under this multi-objective perspective, problems are solved with explicit consideration for several characteristics, potentially conflicting, and solutions can represent different optimal trade-offs between all objectives. As such, the approach can assist users in making informed decisions about complex FL problems by presenting explicit choices where a single-objective approach would yield none. Indeed, these general advantages of multi-objective methods have been recognised across disciplines, and the field of multi-objective optimisation (MOO) has been thriving for decades. %
This success opens another interesting avenue of research in connection with federated learning: deploying FL methods to facilitate multi-objective learning in distribution, where problems would otherwise be difficult to solve for participants that cannot share local training data.\\ 
Recent works in the literature have begun to combine federated learning with MOO methods to address a wide range of challenges. However, the broader context of the intersection between MOO and FL has not yet been discussed. This work aims to provide a first such systematic overview, identifying general challenges and parallels, and formulating a novel taxonomy to classify existing work while highlighting open directions of research. 
Many FL strategies already use (linear) combinations of multiple functions as objectives, but do not consider the problem from a multi-objective angle.
The first works to explicitly introduce multi-objective methods to Federated Learning aimed to improve federated aggregation and introduce fairness between clients~\cite{hu_fedmgda_2020}, followed by approaches introducing other, system-wide aggregation parameters~\cite{mehrabi_towards_2022}. Another early adoption of MOO was in hyperparameter optimisation for FL~\cite{zhu_multi-objective_2019}. More recently, research has also begun into supporting the inverse scenario: developing strategies to federate the solving of multi-objective problems by distributed clients, e.g.~\cite{yang_federated_2023}\cite{hartmann_mofld_2023}. 
The contributions of this work can be summarised thus:\begin{itemize}
    \item We propose a novel taxonomy of algorithms combining MOO methods and FL, offering a unified naming system for works at the intersection of two previously largely separate fields with separate naming conventions.
    \item We present a thorough review of the state of the art, categorising and contrasting existing works.
    \item We highlight open questions and offer perspectives on open avenues for future research.
\end{itemize}
The rest of this work is organised as follows: Section~\ref{sec:background} reviews important notions from the fields of FL and MOO. Section~\ref{sec:taxonomy} introduces our taxonomy, discussing in detail each category and relevant works from the literature. Finally, we offer a conclusion and perspectives on future research in Section~\ref{sec:conclusion}.
\section{Background}\label{sec:background}
In this section, we briefly introduce fundamental concepts from the fields of federated learning and multi-objective optimisation in preparation for the main body of the survey.
\subsection{Federated Learning}  %
\begin{figure}
    \centering
    \includegraphics[width=.7\linewidth]{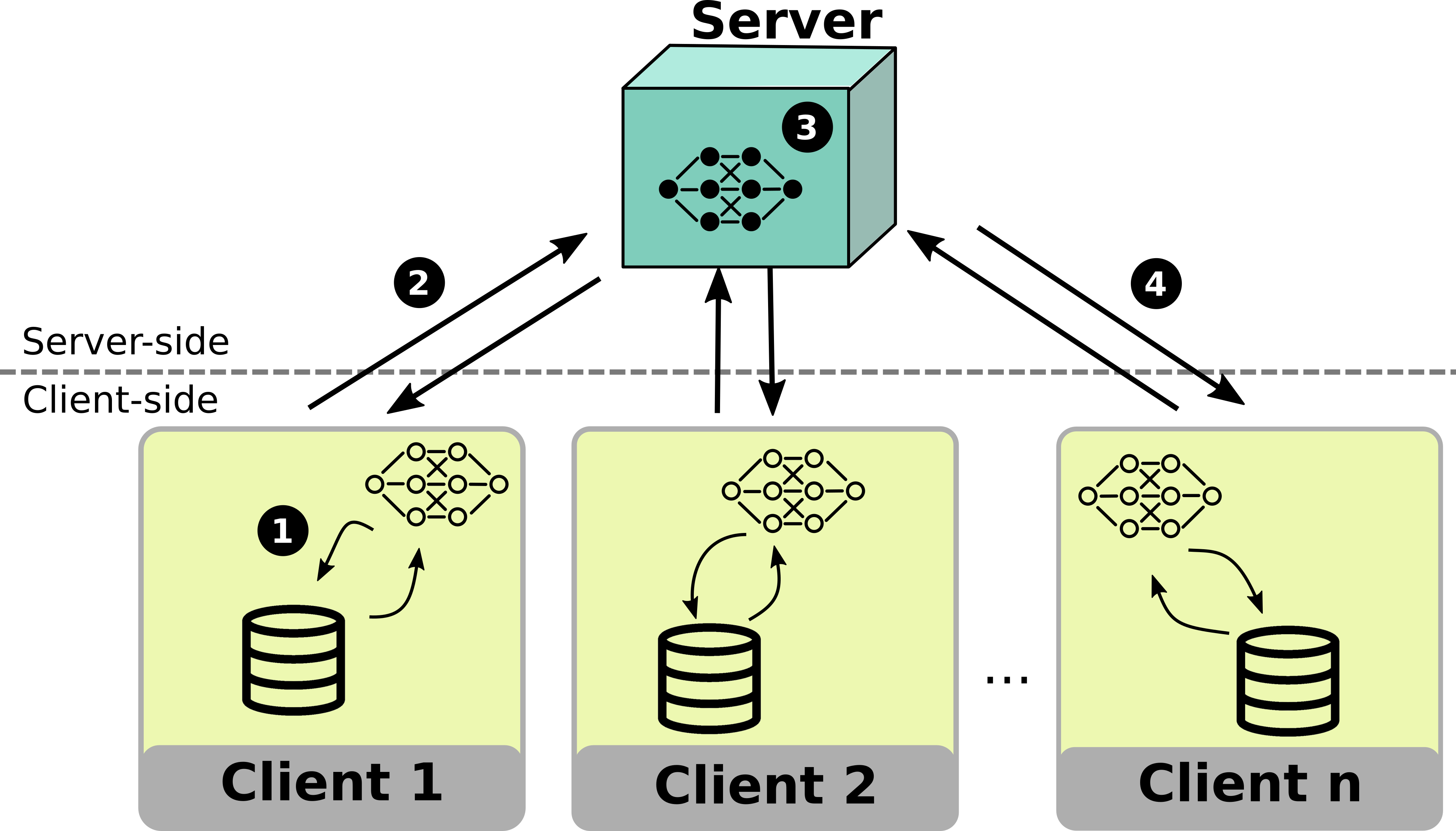}
    \caption{The FL paradigm. During each round, clients perform local model training (1), then transmit their local models to the server (2) for aggregation into a single global model (3). The global model is returned to the clients (4) to begin the next training round.}
    \label{fig:illustration-federated-learning}
\end{figure}

The Federated Learning~\cite{mcmahan_communication_2017} paradigm was originally designed to solve arbitrary (neural network-based) machine learning problems in a difficult distributed setting. This setting is characterised by (i) the available data originating in distribution, with no control over the composition of the resulting datasets, and (ii) a restriction on transmitting private client information, including raw training data, between participants.
FL overcomes the constraint introduced by (ii) by training separate local models in distribution on each dataset holder, or client, and aggregating only the resulting models across clients  -- see Figure~\ref{fig:illustration-federated-learning}.\\
A more detailed general framework of the Federated Learning strategy is presented in Algorithm~\ref{alg:fl-framework}, with colours highlighting the correspondence of code segments to different levels of the federated system (to be presented in detail in Section~\ref{subsec:integrating-mo-fl}). First, the federated system is initialised with the identity of the server, a list of participating clients, and the definition of the underlying learning problem to be solved. Additional hyperparameters are passed depending on the specific algorithm, defining e.g.~the architecture of the neural network to be trained, a client sampling rate, gradient thresholds, or any other parameter required by the algorithm. Then, the local learning process begins.
During each federated training round, a set of clients is selected for participation. These clients each carry out local training and return the resulting models to the server. 
These local models are aggregated periodically by the server into a single global model incorporating the locally learned information. The global model is then passed back to the local clients to continue the next local training round.
Expressed formally, the FL process aims to find a global model $\theta$ that generalises to all available data, i.e.~\begin{equation}
    minimise_{\theta} f(\theta , \mathcal{D}),
\end{equation}
where $\mathcal{D} \defeq \bigcup_i^n \mathcal{D}_i$, with $\mathcal{D}_i$ the dataset of the $i$-th client.
Imbalances between client datasets, as can be caused by characteristic (i), represent a significant challenge to the model aggregation step of FL algorithms. Indeed, any type of heterogeneity between clients, e.g.~in terms of hardware capability or feature distribution, may have an adverse impact on the convergence of the federated model. Mitigating the impact of various types of client heterogeneity remains an active field of study. Other major research topics in FL include the reduction of resource consumption -- mainly computing and communication cost -- and how to protect against malicious actors. For a comprehensive overview of the state of the art in the field, we refer to \cite{kairouz_advances_2021}.
\begin{algorithm}[bt]
    \caption{The general Federated Learning framework.}
    \label{alg:fl-framework}
    \tikzmk{left}\textbf{Input}: Server, list of clients, local learning problem.\\
    \textbf{Parameter}: Optional list of hyperparameters.\\
    \textbf{Output}: Global model $\theta $.
    \begin{algorithmic}[1] %
        \State \colorline{Initialise system parameters.}
        \tikzmk{A}
        \While{stopping condition not satisfied}
        \ForAll{participating clients}
        \tikzmk{B}\boxit{newboxgreen}
        \tikzmk{A}
        \While{local stopping condition not satisfied}
            \State Perform training on local data.
        \EndWhile\tikzmk{B}\boxit{newboxyellow}
        \tikzmk{A}
        \State Transmit local model to server.
        \EndFor
        \State Aggregate local models to obtain new global model.
        \State Return global model to clients.
        \EndWhile\tikzmk{B}\boxit{newboxgreen}
        \State \textbf{return} global model.
    \end{algorithmic}
\end{algorithm}

\subsection{Multi-objective optimisation}  %
\begin{figure}[H]
    \centering
    \includegraphics[width=.34\linewidth]{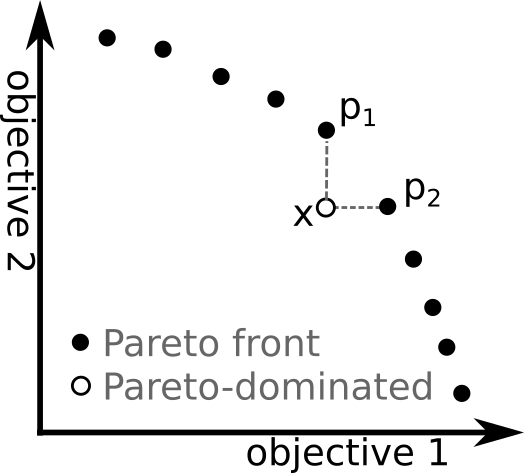}
    \caption{Pareto front and Pareto dominance. Shaded markers represent solutions on the Pareto front of a bi-objective maximisation problem; $x$ is Pareto-dominated by $p_1$ and $p_2$.}
    \label{fig:illustration-pareto-front}
\end{figure}

Multi-objective optimisation is concerned with solving problems in the presence of more than one objective. As an example, consider the problem of selecting hyperparameters for a neural network to simultaneously maximise model utility and minimise the cost of training. Instead of a single objective $f(x)$, such a multi-objective problem is expressed as a vector of $n$ objectives $\vec{f}(x)\defeq(f_1(x), \dots, f_n(x))^T$. Note that individual objectives can conflict, i.e.~in general no single solution can optimise all objectives simultaneously. %
Instead, MOO methods typically focus on identifying solutions that represent an optimal trade-off between objectives, where objective values are balanced so that no single objective can be improved without sacrificing the performance of another. Such trade-off solutions are known as \textit{Pareto-optimal}. Pareto optimality can be difficult to determine in practice, where the optimal values achievable for each objective are unknown, so the weaker notion of \textit{Pareto-dominance} %
is commonly used instead. A solution $x$ is said to Pareto-dominate another solution $y$ iff it outperforms $y$ in at least one objective while matching or improving the value of all others. Formally, \begin{equation}
    x \succ_P y \iff \exists j f_j(x) > f_j(y) \land \forall i f_i(x)\geq f_i(y)
\end{equation} for a maximisation problem. %
Pareto-optimal solutions are not dominated by any others. The set of such solutions is known as the \textit{Pareto front} (see Fig.~\ref{fig:illustration-pareto-front}). %
Most MOO algorithms are either designed to find such a Pareto front, or a single solution based on predefined requirements such as user preferences. A wide range of algorithmic approaches exists for both variants, tailored to different problem characteristics. In this work, we will discuss relevant MOO strategies as they appear; for a comprehensive overview we refer to \cite{talbi_metaheuristics_2009}.
\subsection{Integrating multi-objective methods and Federated Learning}\label{subsec:integrating-mo-fl}
\begin{figure}[t]
    \centering
    \includegraphics[width=.5\linewidth]{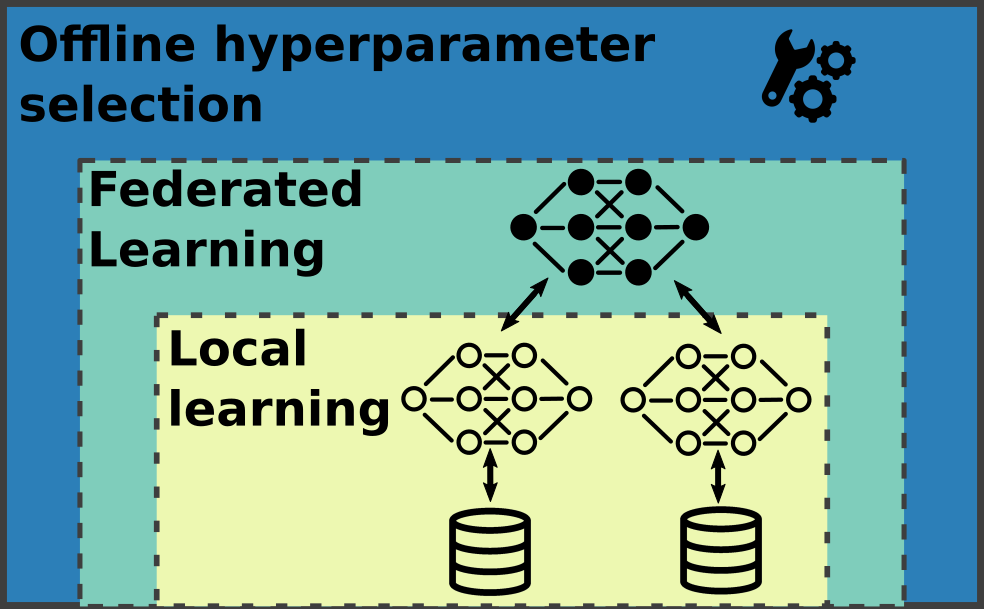}
    \caption{Relation of major categories of the taxonomy. Multi-objective methods can be integrated at different levels of the federated system: in the local learning process of clients, at system-level in the federated algorithm, or outside of the federated system.}%
    \label{fig:taxonomy-boxes}
\end{figure}

We note that multi-objective methods can be integrated with FL at different levels of the federated system, each with distinct implications for the algorithmic components involved. Based on this insight, we propose a three-level view of the federated system -- see Fig.~\ref{fig:taxonomy-boxes} and corresponding colours in Alg.~\ref{alg:fl-framework}.
Adding multi-objective methods on top of a federated algorithm necessitates no modification of the underlying federation or local learning process; an example for such a method is offline hyperparameter tuning with respect to multiple requirements. On the other hand, introducing multi-objectivity at the federated level, e.g.~for model aggregation on the server, forces adaptation at the top level as well: any hyperparameter algorithm running on the federated system must accommodate new parameters introduced by multi-objective methods.
Finally, adding a multi-objective perspective to the lowest level in Fig.~\ref{fig:taxonomy-boxes} -- the client level -- requires modifications across the entire system: \begin{enumerate*}[label=(\roman*)]\item~The local learning algorithm on each client must handle multi-objective problems; \item~the federated algorithm must aggregate client submissions, which may include multi-objective gradients or be influenced by heterogeneous client objectives, and \item any hyperparameter must be adjusted once again.
\end{enumerate*}

\section{Taxonomy: multi-objective methods in FL}\label{sec:taxonomy}
\begin{figure*}
    \centering
    \includegraphics[width=\linewidth]{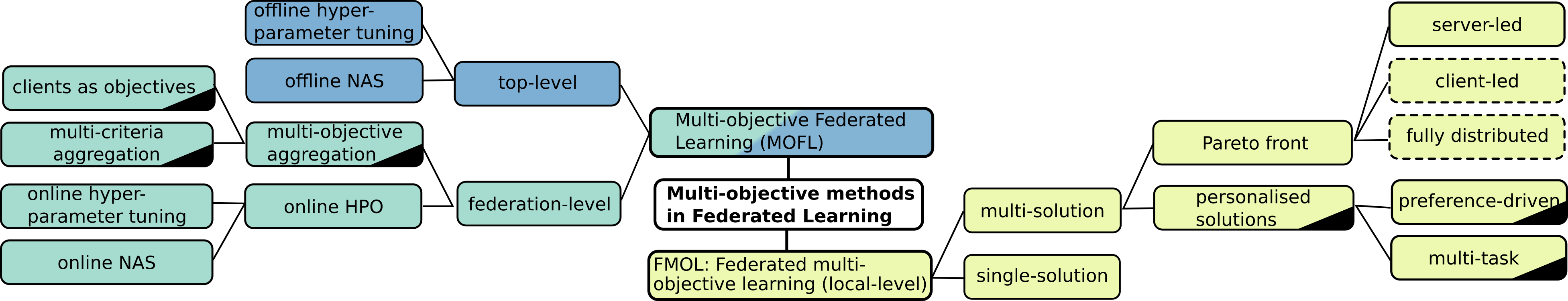}
    \caption{Proposed taxonomy. Colours denote the level of the federated system where MO methods are integrated (see Fig.~\ref{fig:taxonomy-boxes} and Sec.~\ref{subsec:integrating-mo-fl}). Some categories arise from the unique properties of the FL setting; these are marked by a shaded corner. Categories in dashed boxes are currently unexplored in the literature.
    }
    \label{fig:taxonomy-tree}
    \vspace{-1em}
\end{figure*}

In this section, we introduce our proposed taxonomy, discussing each category and the related existing work. The full taxonomy is shown in Figure~\ref{fig:taxonomy-tree}.
A first fundamental distinction is the purpose that multi-objective and federated methods each serve in an algorithm. %
We can identify two main broad categories: one where MOO methods are applied to enhance the functionality of a federated system, and the inverse, where FL is used in support of solving a general multi-objective problem in distribution. We refer to these categories as \textit{Multi-Objective Federated Learning (MOFL)} and \textit{Federated Multi-Objective Learning (FMOL)}, respectively, to indicate the different chaining of strategies. 
MOFL %
covers the majority of existing research, and is notably precisely equivalent to the top two layers as shown in Fig.~\ref{fig:taxonomy-boxes} and introduced in Section~\ref{subsec:integrating-mo-fl}. Works in this section can accordingly be divided further into top-level and federation-level methods, and will be discussed as such in the following sections.
FMOL, in contrast, corresponds to the lowest layer in Fig.~\ref{fig:taxonomy-boxes}, and extends the ``standard'' FL scenario, where Federated Learning is used to solve an arbitrary learning problem in distribution, to include multi-objective learning problems. %
\subsection{Multi-objective federated learning at top level}
Methods at the top level of a federated system, as defined in Fig.~\ref{fig:taxonomy-boxes}, are decoupled from the federated learning and aggregation process and can treat the federated algorithm as a black-box system. As such, this class of algorithms is arguably the least specific to the FL context, since modifications at this level require no particular adaptation to the federated setting. 
Current work can largely be divided into two major applications: multi-objective neural architecture search (MO-NAS), focused on optimising the architecture of a neural network with respect to multiple objectives, and more general multi-objective hyperparameter tuning, where other hyperparameters of the federated system are tuned. Both types typically employ population-based multi-objective strategies, known to offer effective search space exploration.
\subsubsection{Offline hyperparameter tuning}
Multi-objective hyperparameter tuning can find algorithm parameters for additional requirements beyond the utility of the global model. Depending on the use case, FL systems may face challenges such as privacy restrictions, resource limitations, %
or malicious attacks. This approach allows users to explicitly model such requirements and make informed choices about the trade-offs inherent to different solutions.\\
\cite{kang_hyperparameter_2024} assert that optimising hyperparameters %
solely for model performance may expose the federation to a risk of data leakage. The proposed mitigation approach optimises the three objectives of model performance, training cost and privacy leakage simultaneously. This algorithm, derived from NSGA-II~\cite{deb_fast_2002}, a well-known population-based baseline algorithm, is designed to find a Pareto front of possible configurations representing different trade-offs between these objectives. 
\cite{morell_multi-objective_2024} also introduce a second objective in addition to the model accuracy, based on the mean amount of data transmitted and received by clients. This approach is designed to optimise a large number of hyperparameters and algorithmic choices, including the number of local training steps, the number of bits used to encode local updates, and whether clients submit gradient or weight updates. All variables are optimised using a hybrid of NSGA-II and an estimated distribution-based algorithm (EDA). 
\cite{geng_multiobjective_2024} formulate a similar strategy, also using NSGA-II, but considering the four objectives of minimising global model error rate, the variance of model accuracy, the communication cost, and a privacy budget. %
\subsubsection{Offline neural architecture search}
Neural architecture search aims to optimise the structure of a neural network for given objectives. Federated NAS can be seen as an inherently multi-objective problem~\cite{zhu_federated_2021}, as changes to the model structure impact not only the model utility, but also other aspects of the federated system, such as the communication and training cost.
One of the first works on multi-objective federated neural architecture search~\cite{zhu_multi-objective_2019} proposes an offline federated NAS algorithm that constructs models with the two objectives of minimising the validation error obtained by the model, and the cost of communicating the model. Solutions are once again generated using NSGA-II. %
The same problem is tackled in \cite{chai_communication_2023}, but with the use of a multi-objective evolutionary algorithm (MOEA) instead of NSGA-II to improve the exploration of the multi-objective search space.\\
Federated split learning is a related problem, where partial blocks of the global model are assigned to clients, with blocks of different size assigned to clients depending on the available resources.~\cite{yin_predictive_2023} propose to optimise this splitting decision, along with communication bandwidth and computing resource allocation, as a multi-objective problem, minimising training time and energy consumption of the system. The proposed algorithm yields a Pareto front of solutions using a hybrid of NSGA-III and a generative adversarial network trained to identify configurations generating Pareto-dominated solutions.
Research on offline MO-NAS algorithms for FL is arguably more advanced than other areas of MOFL, as existing approaches can be applied to the federated setting without change. The main challenge remains the high computational cost of these methods.

\subsection{Multi-objective federated learning at federation-level}

\begin{table*}[t]
    \centering
    \begin{tabularx}{\textwidth}{lZZZZ}
        \toprule
        Reference & Taxonomy label & System level & MOO method & Objectives \\
        \midrule
        \cite{hu_fedmgda_2020} & Clients as objectives & federation-level & MGDA & Local model utilities \\
        \cite{ju_accelerating_2024} & Clients as objectives & federation-level & dynamic preferences & Fairness, convergence \\
        \cite{mehrabi_towards_2022} & Multi-criteria aggregation & federation-level & obj.-contribution scoring & Arbitrary system objs. \\
        \cite{zhu_realtime_2022} & online MO-NAS & federation-level & NSGA-II & Global model utility, evaluation speed \\
        \cite{kang_hyperparameter_2024} & offline MO-HPO & top-level & NSGA-II & Model utility, training cost, privacy leakage\\
        \bottomrule
    \end{tabularx}
    \caption{Comparison of selected MO-FL algorithms. Each row lists the level of the federated system where multi-objective notions are introduced, as well as the method used to solve the multi-objective problem.}
    \label{tab:mo-fl-algorithms}
\end{table*}

MOO methods can also be integrated with FL at the server-level to solve challenges inherent to the FL paradigm -- a brief overview of representative works from the literature is presented in Table~\ref{tab:mo-fl-algorithms}. The majority of existing works focus on one of two design aspects of a federated system: the aggregation strategy used on the federated server, and the selection of relevant hyperparameters for the FL algorithm. We discuss both separately, beginning with multi-objective aggregation.
\subsubsection{Multi-objective aggregation} The aggregation of local model updates by the server can be modelled as a MOO problem, permitting the use of more than one criterion for computing the global model. %
This multi-objective version of federated aggregation can be formulated in general terms as follows: \begin{equation}\label{eq:mo-aggregation-problem}
min_\theta (f_1(\theta),\dots, f_n(\theta)),^T
\end{equation}
where $\theta$ is the global model and $f_i$ is the loss function of the $i$-th objective. Solving this problem typically translates to finding optimal aggregation weights $\lambda_i$ to compute the global model from the local models: \begin{equation}
    min_{\lambda_1\dots,\lambda_n} (f_1(\theta),\dots, f_n(\theta)),^T,~\text{with }\theta = \sum_i^n \lambda_i\theta_i.
\end{equation}
The literature on FL algorithms with multi-objective aggregation can be categorised based on the nature of the objectives~\cite{kang_optimizing_2024}. One line of work derives objectives from the performance of individual clients; the other uses objectives that describe the federation as a whole. This distinction is significant, as the different mathematical properties of these variants %
permit the use of different multi-objective methods. The following sections discuss both types in detail.
\paragraph{Clients as objectives.} These algorithms consider the performance of individual clients and the global model as separate objectives. In client-heterogeneous settings, this approach can balance the interests of both the clients and the general system. This perspective enables explicit fairness guarantees for selfish participants, ensuring that the performance of individual clients is not sacrificed for that of the system in computing the global model. %
Crucially, performance criteria in this class of MOFL problems are tied directly to the client models and thus differentiable with respect to model parameters. As such, they can be solved efficiently using gradient-based multi-objective algorithms such as the classical multi-gradient descent algorithm (MDGA)~\cite{desideri_multiple_2012}, established in the field of MOO.\\
The FedMGDA+ algorithm~\cite{hu_fedmgda_2020} leverages this insight, defining the performance of each participating client as a separate objective. %
Using MGDA yields aggregation weights for a common descent gradient for all clients, thus guaranteeing that no client suffers a reduced performance by participating in an aggregation step. An added constraint on the divergence of aggregation weights serves as protection against false updates by malicious participants.
The FedMC+ algorithm~\cite{shen_multi-objective_2024} is also designed to reconcile individual client updates and the global model in the presence of heterogeneous data. A secondary objective, minimising conflict between the global and local gradients, is introduced during the aggregation step %
and solved by transformation into a convex optimisation problem. %
\cite{cui_addressing_2021} formulate the aggregation step as a parameterised min-max optimisation problem. Fairness constraints serve to optimise model utility for the single worst-performing client while ensuring that \begin{enumerate*}[label=(\roman*)]\item the utility of all clients improves, and \item no client improves much less than another.\end{enumerate*} The solution obtained from this formulation is optimised further to guarantee Pareto-stationarity, a prerequisite for local optimality~\cite{ye_pareto_2022}.\\
The three methods have different implications for the ultimate balance of client models. While both \cite{hu_fedmgda_2020} and \cite{cui_addressing_2021} (in its pure form) guarantee that all clients improve during an aggregation step, only the latter considers the magnitude of gradients in the calculation. Thus, \cite{cui_addressing_2021} may force a greater balance between clients, to the potential detriment of overall performance in highly heterogeneous settings. In contrast, \cite{shen_multi-objective_2024} may sacrifice an outlier for the benefit of the system. %
Though undesirable to selfish clients, the latter could offer a defence against intentionally divergent updates submitted by a malicious client.
\paragraph{Multi-criteria aggregation.} These algorithms perform aggregation based on multiple metrics that describe different characteristics of the federated system, such as the accuracy of the global model and fairness between clients. Such criteria are not generally differentiable with respect to the model, and thus cannot be optimised using gradient-based methods~\cite{kang_optimizing_2024}. Solution approaches rely instead on heuristic insights or the formulation of the aggregation step into a mathematically solvable optimisation problem.\\
\cite{mehrabi_towards_2022} propose an algorithm that can incorporate multiple arbitrary system objectives, including fairness metrics, on the server. Aggregation is accomplished by assigning weighted ranking scores to each client for its contribution to optimising each objective, calculated using a validation dataset possessed by the server. These scores are used to compute aggregation weights.
In contrast,~\cite{ju_accelerating_2024} formulate fairness-controlled FL as a dynamic multi-objective problem, where the optimisation problem consists of a linear combination of client losses, with weights adjusted dynamically to balance the progress of all component objectives. %
This approach yields different trade-off solutions between fairness and convergence depending on the value chosen for a fairness parameter. The idea of optimising a weighted linear combination of objectives in the federated aggregation step was proposed before in \cite{li_fair_2020}, generalising ideas from \cite{mohri_agnostic_2019}; but neither work explicitly acknowledges a multi-objective view of the problem. %
Both aggregation strategies have different strengths and weaknesses. \cite{mehrabi_towards_2022} offers transparent server-side evaluation of clients, including the potential to automatically recognise low-quality or malicious clients. However, the need for a validation dataset on the server may violate the privacy requirements of clients, and renders the method vulnerable to data poisoning attacks. Conversely, \cite{ju_accelerating_2024} offers mathematical fairness guarantees, but little transparency in the aggregation process. In addition, this algorithm may be vulnerable to malicious client participation. %
\subsubsection{Online multi-objective hyperparameter optimisation}\label{subsubsec:mo-meta}
Algorithms that use MOO to optimise hyperparameters for the federated system may run off-line or on-line. In on-line algorithms, the optimisation process is integrated into the federated algorithm, i.e.~parameters are changed during the runtime of the FL process. On-line candidate generation is typically integrated on the federated server at the aggregation step, with local training rounds used for evaluation. %
Existing works on online MO-HPO in FL can again be divided into hyperparameter tuning and neural architecture search.
\paragraph{Online hyperparameter tuning.} 
The work by \cite{badar_fairtrade_2024} performs on-line hyperparameter optimisation for clients, generating and transmitting new parameters during each aggregation step. These parameters, a fairness constraint regularisation parameter and the learning rate designed to enforce fairness locally, are recomputed on the server-side by using multi-objective Bayesian optimisation.
Finally,~\cite{banerjee_optimized_2022} propose a multi-objective on-line device selection approach to speed up the learning process in the presence of stragglers. The selection algorithm is designed to maximise the available computing and communication resources on selected clients, using NSGA-II. %
\paragraph{Online neural architecture search.} 
NAS algorithms may be designed run on-line, modifying during the execution of the federated algorithm the structure of the neural network to be trained by each client. Such a strategy could significantly reduce the computational cost of the search, at the price of complicating the training and aggregation process by introducing dynamic parameters. %
The only such algorithm currently existing in the MOFL literature dynamically optimises the accuracy and evaluation speed of federated model training~\cite{zhu_realtime_2022}. The NSGA-II algorithm is used during each aggregation step to generate partial samples of the full model to assign to clients for training.
On-line MO-NAS presents a difficult challenge and is currently underexplored in the literature, but could offer significant efficiency benefits.

\subsection{Federated multi-objective learning}
\begin{table*}[t]
    \centering
    \begin{tabularx}{\textwidth}{lZZZZ}
        \toprule
        Reference & Taxonomy label & Local MOO method & Global MOO method & Objectives \\
        \midrule
        \cite{yang_federated_2023} & single-solution & successive single-obj.~updates & MGDA & arbitrary \\
        \cite{askin_federated_2024} & single-solution & linearised objectives & MGDA & arbitrary \\
        \cite{hartmann_mofld_2023} & server-led & linearised objectives & offline metaheuristic & arbitrary \\
        \cite{sen_fedmtl_2024} & multi-task & multi-task layer & similarity-based partial aggregation & arbitrary separable tasks \\
        \cite{hartmann_fedpref_2024} & preference-driven & linearised preferences & similarity-based aggregation+clustering & arbitrary\\
        \bottomrule
    \end{tabularx}
    \caption{Comparison of selected Federated Multi-objective Learning (FMOL) algorithms. Note that all algorithms are dedicated to handling local multi-objective learning. As noted in Section~\ref{subsec:integrating-mo-fl}, this requires modifications at several levels of the federated system.}
    \label{tab:fmol-algorithms}
\end{table*}
In federated multi-objective learning, the solving of a multi-objective learning problem (MOLP) is the ultimate goal, and FL acts as an auxiliary tool to facilitate learning in distribution. A major challenge compared to the class of MOFL algorithms is that in this setting, there is no control or information about the compatibility of the objectives involved in the problem, whereas in MOFL the objectives were designed to suit the federated setting. Note also that FL techniques have largely been developed for neural networks, so the focus in this setting is on MO-algorithms that train such models.
Compared with the application of MO techniques to FL algorithms, the federated solving of MOLPs has received very little attention so far. Here we aim to offer a classification of the few existing works, and extrapolate the open challenges and problems that remain to be solved. See also Table~\ref{tab:fmol-algorithms} for a representative overview of existing works. 
On the most fundamental level, algorithms in this category can be separated by the number of solutions they are designed to find: one single solution to the MOLP, or multiple solutions representing different trade-offs between the underlying objectives. 
\subsubsection{Methods finding a single solution}
FMOL algorithms designed to find a single solution aim to find an arbitrary Pareto-stationary solution. The advantage of such approaches is a relatively quick convergence, e.g.~by exploiting gradients to locate the nearest solution. The main disadvantage is a lack of control over which solution out of all possible ones is found, and thus a lack of choice for potential users. 
One of the earliest such works~\cite{yang_federated_2023} once again extends the MGDA algorithm to the federated setting, this time with respect to client objectives. Local training sequentially updates client models with respect to each component objective. Then, clients submit a gradient vector for aggregation to the server, where MGDA yields optimal aggregation weights to update the global model. %
This algorithm is shown to converge to a Pareto-stationary solution.
A subsequent work~\cite{askin_federated_2024} points out a risk of local drift in this approach, as well as a high communication load caused by transmitting separate gradient updates for all objectives. The algorithm proposed to mitigate these issues is also based on server-side MGDA, but clients reduce communication cost by transmitting a compressed matrix of all objective gradients. %
Local drift is avoided via a similar modification: client updates are computed from a linear combination of all objective gradients rather than a series of single-objective updates. Tackling a different use case,~\cite{kinoshita_federated_2024} discuss data-driven MOO problems, where a federated server attempts to solve a multi-objective problem, e.g.~clustering, using only indirect information from distributed clients. %
In this unsupervised setting, no gradient-based strategies are possible; the server instead utilises a MOEA to solve the problem.
\subsubsection{Methods finding multiple solutions}
Federated algorithms designed to find multiple solutions have one of two goals: they either attempt (1) to find a full Pareto front, i.e.~a set of trade-off solutions, or (2) to find a personalised model for each participant. For both variants, participants may have different preferences over the same objective functions, or may even be solving entirely disjoint tasks.
\paragraph{Finding a Pareto front.} Algorithms that aim to find a Pareto front of solutions must explore a wide range of the search space to identify a diverse spread of trade-off solutions. In the distributed setting, this may happen at different levels of the federated system: \textit{server-led} exploration sees the federated server managing the exploration and constructing a Pareto front. A first framework for such a scenario has been proposed in \cite{hartmann_mofld_2023}, utilising a metaheuristic on the federated server to decompose the multi-objective problem in into single-objective candidate subproblems. This approach bears similarities to some of the top-level algorithms discussed in Section~\ref{subsubsec:mo-meta}, in that each candidate is evaluated separately by a full federated system. Unlike those approaches, however, the full system is not strictly required for an effective evaluation. %
Thus, the efficiency of the evaluation could be improved by the use of an algorithm that can federate candidates with different objective preferences. To the best of our knowledge, such an algorithm has not yet been proposed in the literature. Future contributions may be able to leverage client-specific solution algorithms in combination with server-led Pareto exploration strategies.\\
In contrast, \textit{client-led} exploration would have each client attempting to find a Pareto front, e.g.~in cases where the server is untrusted or lacks computing resources. This scenario has, to the best of our knowledge, not yet been addressed in the literature, but would carry its own challenges and opportunities inherent to the federated setting, most importantly a shift of control from server to clients, and the alignment of local Pareto fronts. Possibly related is the \textit{fully-distributed} setting, where no server is involved in the training process and aggregation is decentralised across the client network.
\paragraph{Finding client-specific solutions.} 
Here, the goal of the algorithm is to find a solution for each client in the system, based on different local requirements. Crucially, and in contrast to single-solution algorithms, this approach yields a different model for each client, matching that client's objectives, instead of finding a global model that generalises over all clients. This variant is known as Personalised FL, and is typically used in highly heterogeneous settings where the focus is on individual client performance~\cite{tan_towards_2023}. Note that this type of algorithm is arguably unique to the federated setting, arising from its properties that participants in FL are heterogeneous and may have different, independent interests.

In a \textit{preference-driven} setting, client heterogeneity is induced by different preference weights assigned by each client to the same underlying multi-objective problem \cite{hartmann_fedpref_2024}. %
Formally, the objectives of the $i$-th client are weighted by that client's unique preference weights $w^i$: \begin{equation}
    \vec{f}^i(x) \defeq \vec{w}^i\odot \vec{f}(x) = (w^i_1f_1(x),\dots, w^i_nf_n(x))^T %
\end{equation}
Where objective components are conflicting, learning trajectories of clients could diverge even on the same underlying model; the PFL approach is intended to embrace this diversity instead of counteracting it.
Only a handful of works so far have considered a personalised approach to objective heterogeneity.
In the first such work~\cite{hartmann_fedpref_2024}, client preferences are assumed to be private, and local training is performed on a  weighted linear combination of the objectives. The challenge in this setting is to aggregate clients whose current training trajectory is compatible, and separate clients where it is not. As little direct information about the mutual compatibility of clients is available on the server, many classical MOO methods cannot be applied. Instead, the proposed algorithm performs clustering and weighted aggregation based on the similarity of model updates.\\
Federated \textit{multi-task} learning is an edge case scenario where clients solve mutually different subsets of tasks (i.e.~objectives). A number of works in the FL literature, e.g.~\cite{ghosh_efficient_2020} and \cite{huang_federated_2023}, have addressed a simplified setting where each client is assigned a single task\footnote{Note that the `multi-task' label is assigned inconsistently in the existing FL literature, referring variously to clients with heterogeneous datasets or objectives.} without acknowledging a multi-objective perspective. 
To the best of our knowledge, only one work currently considers the problem where each client is assigned a set of several tasks~\cite{sen_fedmtl_2024}. Similarly to other works on FMOL, this task assignment is private. Under the proposed algorithm, clients jointly train a block of shared model parameters plus a separate parallel model layer for each task to be solved by the client. Once again, clients are aggregated based on a model similarity score, computed here based both on the shared parameters and a matching of task-specific layers.

\section{Conclusion and perspectives}\label{sec:conclusion}
In this work, we have presented the first comprehensive survey on the use of multi-objective methods in connection with Federated Learning. We have proposed a novel taxonomy to classify existing works in the literature, and offered a perspective on recent trends, open challenges and possible approaches.
 Existing work demonstrates that MOO is a promising tool to improve transparency and effectiveness of FL techniques when navigating real-world problems. As in the wider field of FL, further work remains to be done. %
Open avenues of research in MO-FL include, most prominently, \begin{enumerate*}[label=(\roman*)]\item effective defence against malicious attackers in multi-objective aggregation; \item the use of MOO methods specifically to recognise low-quality clients; \item enhancing transparency and control of MO-preferences for users, e.g.~by generating multiple different Pareto-optimal solutions, and \item exploring more sophisticated MOO techniques, e.g.~to replace the baseline NSGA-II algorithm that is currently used in many of the works discussed here. The area of FMOL, enabling the federated solving of multi-objective learning problems, remains largely open. Initial contributions to the field could include, for example, \item improving the efficiency of server-led strategies finding a Pareto front; \item exploring the effect of preference heterogeneity on convergence in single- and multi-solution algorithms; \item exploring the cumulative effect of data heterogeneity on FMOL problems; \item considering variant FMOL settings, e.g. where client preferences are not private. \end{enumerate*}
\newpage
\bibliographystyle{named}
\bibliography{references}
\end{document}